\documentclass[11pt, journal, twocolumn]{IEEEtran}

\usepackage[letterpaper, left=1in, right=1in, bottom=1in, top=1in]{geometry}

\usepackage{xcolor,soul,framed} 

\colorlet{shadecolor}{yellow}
\usepackage[pdftex]{graphicx}
\graphicspath{{../pdf/}{../jpeg/}}
\DeclareGraphicsExtensions{.pdf,.jpeg,.png}

\usepackage[cmex10]{amsmath}
\usepackage{array}
\usepackage{mdwmath}
\usepackage{mdwtab}
\usepackage{eqparbox}
\usepackage{url}
\usepackage{listings}
\usepackage{xcolor}
\usepackage{booktabs}
\usepackage{tabularx}
\usepackage{array}
\usepackage[flushleft]{threeparttable}
\usepackage{threeparttablex}
\usepackage{hyperref} 
\usepackage{lipsum}
\usepackage[ruled,vlined,linesnumbered]{algorithm2e}
\usepackage{footnote}
\usepackage{caption}
\usepackage{subcaption}
\usepackage{amsmath}
\usepackage{amssymb}
\usepackage{stmaryrd}

\usepackage{lineno}
\usepackage{easylist}
\usepackage{amssymb}
\usepackage{amsmath}
\usepackage{subcaption}
\usepackage{url}
\usepackage{textcomp}
\usepackage{verbatim}
\usepackage[ruled,vlined]{algorithm2e}
\usepackage{ulem}
\usepackage{graphicx}
\usepackage{pgfplots}
\usepackage{listings} 
\pgfplotsset{compat=1.14}
\pgfplotsset{compat=newest}
\pgfplotsset{plot coordinates/math parser=false}
\usepackage{tikzscale}
\usetikzlibrary{matrix,chains,positioning,decorations.pathreplacing,arrows}
\usepackage{tikz-qtree,tikz-qtree-compat}
\usetikzlibrary{calc}
\modulolinenumbers[5]

\usepackage{caption}

\newcommand{\node}{t}
\newcommand{\bx}{\mathbf{x}}

\newcommand{\R}{\mathbb{R}}

\newcommand{\cI}{\mathcal{I}}

\newcommand{\lc}{\textrm{left}}
\newcommand{\rc}{\textrm{right}}

\newcommand{\Impurity}{\textrm{I}}
\newcommand{\decrease}{\Delta_{\cI}}
\newcommand{\tree}{T}

\lstdefinestyle{pythonstyle}{
    commentstyle=\color{codegreen},
    keywordstyle=\color{magenta},
    numberstyle=\tiny\color{codegray},
    stringstyle=\color{codepurple},
    basicstyle=\ttfamily\footnotesize,
    breakatwhitespace=false,         
    breaklines=true,                 
    captionpos=b,                    
    keepspaces=true,                 
    numbers=left,                    
    numbersep=5pt,                  
    showspaces=false,                
    showstringspaces=false,
    showtabs=false,                  
    tabsize=2
}

\begin{document}

\renewcommand\IEEEkeywordsname{Keywords} 

\bstctlcite{IEEEexample:BSTcontrol}
  

\title{Federated unsupervised random forest for privacy-preserving patient stratification}  


\DeclareRobustCommand*{\IEEEauthorrefmark}[1]{%
  \raisebox{0pt}[0pt][0pt]{\textsuperscript{\footnotesize\ensuremath{#1}}}}  

\author{\IEEEauthorblockN{Bastian Pfeifer\IEEEauthorrefmark{1}$^{*}$}, 
Christel Sirocchi\IEEEauthorrefmark{2}, 
Marcus D. Bloice\IEEEauthorrefmark{1}, \\
Markus Kreuzthaler\IEEEauthorrefmark{1} and
Martin Urschler\IEEEauthorrefmark{1}

\IEEEauthorblockA{
\vspace{0.3cm}
\IEEEauthorrefmark{1}Institute for Medical Informatics, Statistics and Documentation \\ Medical University Graz, Austria \\
\IEEEauthorrefmark{2}Department of Pure and Applied Sciences, University of Urbino, Italy \\
}

\thanks{ 
$^{*}$Corresponding author: Bastian Pfeifer (bastian.pfeifer@medunigraz.at).
}}


\markboth{}{}

\maketitle

\begin{abstract}
In the realm of precision medicine, effective patient stratification and disease subtyping demand innovative methodologies tailored for multi-omics data. Clustering techniques applied to multi-omics data have become instrumental in
identifying distinct subgroups of patients, enabling a finer-grained
understanding of disease variability. This work establishes a powerful framework for advancing precision medicine through unsupervised random-forest-based clustering and federated computing. 
We introduce a novel multi-omics clustering approach utilizing unsupervised random-forests. 
The unsupervised nature of the random forest enables the determination of cluster-specific feature importance, unraveling key molecular contributors to distinct patient groups.
Moreover, our methodology is designed for federated execution, a crucial aspect in the medical domain where privacy concerns are paramount. 
We have validated our approach on machine learning benchmark data sets as well as on cancer data from The Cancer Genome Atlas (TCGA). Our method is competitive with the state-of-the-art in terms of disease subtyping, but at the same time substantially improves the cluster interpretability. Experiments indicate that local clustering performance can be improved through federated computing. \\
\end{abstract}



%
\IEEEpeerreviewmaketitle


\section{Introduction} 
Federated machine learning is crucial in preserving data privacy and security while enabling collaborative model training across decentralized devices or organizations \cite{brauneck2023federated}. It empowers various stakeholders to jointly improve machine learning models without sharing sensitive data, making it ideal for applications in healthcare, where data confidentiality is paramount \cite{dayan2021federated}. 

Disease subtyping has emerged as a pivotal strategy in precision medicine \cite{leng2022benchmark}, aiming to unravel the inherent heterogeneity within complex disorders. Multi-omics data, encompassing genomics, transcriptomics, proteomics, and more, provides a holistic view of biological systems, offering unprecedented insights into the molecular underpinnings of diseases \cite{lipkova2022artificial}. Clustering techniques applied to multi-omics data have become instrumental in identifying distinct subgroups of patients, enabling a finer-grained understanding of disease variability \cite{rappoport2018multi}. These approaches hold immense promise for tailoring treatments and interventions to the specific molecular signatures associated with each disease subtype, thereby advancing the era of personalized and targeted therapeutics. 

This paper introduces a multi-omics clustering approach utilizing unsupervised random-forest to address these challenges. The proposed random forest incorporates a novel unsupervised splitting rule for disease subtype discovery. At the same time, it allows the determination of cluster-specific feature importance, unraveling key molecular contributors to distinct patient groups. Moreover, our methodology is designed for federated execution, a crucial aspect in the medical domain where data privacy is an ethical necessity. 
To the best of our knowledge, there is currently no existing approach designed for federated multi-omics clustering within the context of disease subtype discovery. 

The manuscript is organized as follows: We start with a brief discussion of related work in Section \ref{sec:related_work}. In Section \ref{sec:approach} we describe our approach to federated patient stratification and we mathematically formulate the underlying algorithms. The evaluation strategy is described in Section \ref{sec:evaluation} and is followed by a discussion of the obtained results in Section \ref{sec:results}. We conclude with Section \ref{sec:conclusion}.

\section{Related work} \label{sec:related_work}
The recent years have seen a wide range of advanced methods for multi-omics clustering \cite{subramanian2020multi}. Most prominent is SNF (\textit{Similar Network Fusion}) \cite{wang2014similarity}. For each data (omics) type it models the similarity between patients as a network and then fuses these networks via an interchanging diffusion process. Spectral clustering is applied to the fused network to infer the final cluster assignments. A method which builds upon SNF is called NEMO and was recently introduced in \cite{rappoport2019nemo}. They provide solutions to partial data involving missing values and implement a novel \textit{eigen-gap} method \cite{von2007tutorial} to infer the optimal number of clusters.
\cite{yang2021subtype} proposed a deep adversarial learning method for disease subtyping called Subtype-GAN. Utilizing latent variables obtained from the neural network, Subtype-GAN employs consensus clustering and a Gaussian Mixture Model to discern the molecular subtypes of tumor samples. \cite{yang2023mrgcn} developed a method called MRGCN, which employs a graph convolutional network model for disease subtyping. It concurrently captures and reconstructs the expression and similarity relationships from multiple omics in a unified latent embedding space.


A conceptually related 
method to fuse omics data was introduced by \cite{pfeifer2021hierarchical}. The underlying late-fusion algorithm expects a set of binary matrices, reflecting the clustering solution of each omics type, of dimension $n~\times~n$, where a matrix entry of one means that two samples are in the same cluster, and where $n$ is the number of samples. Based on these matrices the algorithm executes a standard hierarchical bottom-up procedure for building a dendrogram. At each fusion step, the algorithm can use the distances from the available binary views to fuse the samples. The final affinity of two samples is reflected by the number of iterations appearing within the same cluster while building the dendrogram. The aforementioned method was later utilized for ensemble-based multi-omics clustering \cite{pfeifer2023parea}. The authors introduce a method called Parea allowing for layer-wise stacking of ensembles. The Parea ensemble strategy is not constrained to any specific clustering technique. At each layer the clustering strategy and the number of clusters is inferred by an evolutionary algorithm. For more detail we refer to \cite{pfeifer2023parea}. 

However, to the best of our knowledge, no approach has been developed for federated multi-omics clustering in the context of disease subtype discovery.
Federated computing for random forest-based clustering is fairly straightforward and more interpretable compared to other approaches like SNF \cite{wang2014similarity} and NEMO \cite{rappoport2019nemo}. The nature of random forests, which build an ensemble of decision trees independently, aligns well with the decentralized structure of federated learning.

Here, we propose federated unsupervised Random Forest for multi-omics clustering. We introduce a novel unsupervised splitting rule, and we showcase the ability of our method to provide cluster-specific feature importances. 



\section{Materials and methods} \label{sec:approach}
\subsection{Random Forest classifier}
Random Forests \cite{Breiman2001-rl} consist of an ensemble of classification or regression trees. Each individual tree, denoted as $\tree$, represents a mapping from the feature space to the response variable. 
The trees are constructed independently of each other using a bootstrapped or subsampled dataset.

In a tree $\tree$, any given node $\node$ corresponds to a subset, typically a hyper-rectangle, in the feature space. A split of node $\node$ divides the hyper-rectangle $R_\node$ into two separate hyper-rectangles which correspond to the left child $\node^\lc$ and right child $\node^\rc$ of node $\node$, respectively.
For a specific node $\node$ in tree $\tree$, the notation $N(\node)$ 
denotes the number of samples that fall into the hyper-rectangle $R_\node$ and 
\begin{equation}
\label{eq:mun}
\hat{\mathbb{E}}_{\node}\{y\} := \frac{1}{N(\node)}\sum_{i:\bx_i \in R_\node} y_i
\end{equation}
denotes their average response. 
Each tree $\tree$ is grown using a recursive procedure which proceeds in two steps for each node $\node$. First, a subset $\mathcal{M} \subset [p]$ of features is chosen uniformly at random. Then the optimal split variable $x(\node) \in \mathcal{M}$ and split value $z(\node) \in \R$ are determined by maximizing:
\begin{equation*}
\decrease(\node, x(\node), z(\node)) :=    
\end{equation*}
\begin{equation}
\label{eq:impdecrease}
     \Impurity(\node) - \frac{N(\node^\lc) \Impurity(\node^\lc) - N(\node^\rc) \Impurity(\node^\rc)}{N(\node)}
\end{equation}
for some impurity measure $\Impurity(\node)$, typically chosen as MSE,  Gini index, or entropy, where in supervised trees $y$ serves as the response.
We typically refer to $\decrease(\node, x(\node), z(\node))$ as the {\it decrease in impurity} for node $\node$ due to $x(\node)$ and $z(\node)$.


\subsection{A novel unsupervised splitting rule for random forests}

In a random forest employing an unsupervised splitting rule the response vector $y$ is not required to build a tree. Here, we propose an unsupervised splitting rule defined as

\begin{equation}
    \Delta_{\mathcal{F}}(t, x(t), z(t)):=\frac{(D_{\square}(t^{\lc}) + D_{\square}(t^{\rc}))/2 }{ D_{\nabla}(t)}
\end{equation}
where
\begin{equation}
    D_{\square}(t):= \frac{\sum_{i,j} (x_{i}(t)-x_{j}(t))^{2}}{N(\node)(N(\node)-1)}, \quad \text{for} \quad i\neq j
\end{equation}
and 
\begin{equation}
    D_{\nabla}(t):= \frac{\sum_{i,j} (x_{i}(t^{\lc})-x_{j}(t^{\rc}))^{2}}{ N(\node^\lc)N(\node^\rc)},
\end{equation}

where $x$ is a given candidate feature and $x_{i},x_{j}$ refers to the specific value in sample $i$ and sample $j$.
The above specified unsupervised splitting rule is also known as the Fixation Index; often utilized in the field of population genomics to infer population structure \cite{wright1949genetical, holsinger2009genetics}. Essentially, it computes the average pairwise distances between the samples within two groups formed by a node split, divided by the average pairwise distances of the samples between the groups. 

Once the unsupervised random forest is built we count the number of times two samples appear in the same leaf node. 
From these counts an affinity matrix is derived which serves as an input for a hierarchical clustering algorithm. 

In mathematical terms, for each tree \(T_i\), let \(C_i\) be an \(n \times n\) binary matrix, with \(n\) the number of samples. The value of \(C_i[j, k]\) is set to 1, when two samples \(j\) and \(k\) end up in the same leaf node of \(T_i\). After training \(T\) trees, the total (non-federated) count matrix \(C_{\text{local}}\) can be defined as the sum of the binary count matrices from all trees:
\begin{equation}
C_{\text{local}} = \sum_{i=1}^{T} C_i     
\end{equation}

We normalize the total count matrix \(C_{\text{local}}\) by dividing it by its maximum entry:
\begin{equation}
\hat{A}_{\text{local}} = \frac{C_{\text{local}}}{\max(C_{\text{local}})}     
\end{equation}


The affinity matrix $\hat{A}_{\text{total}}$ can be clustered by any distance-based clustering method. Here, we employ Ward's linkage method \cite{ward1963hierarchical,murtagh2014ward} throughout the manuscript. 

In case of multi-omics data the aforementioned procedure is executed for each omics type and we calculate the element-wise sum of the derived count matrices. The resulting count matrix is again normalized by its maximum value. 

\subsection{Federated ensemble learning}
The herein proposed ensemble classifier can be efficiently learned in a federated manner. Same as for federated supervised random forests \cite{hauschild2022federated}, each participant trains a local ''clusterer'' comprising multiple decision trees. The trained decision trees, including their split rules, are then shared by the participants. The federated global model is simply a concatenation of all local trees.

Let \(X^{(k)}\) represent the local client-specific data for client \(k\), and \(M_{\text{global}}\) be the global federated model obtained by concatenating trees from all clients.

\[ M_{\text{global}} = [T_{1}^{(1)}, T_{2}^{(1)}, \ldots, T_{N_1}^{(1)}, T_{1}^{(2)}, T_{2}^{(2)}, \ldots, T_{N_2}^{(2)}, \]
\begin{equation}
\ldots, T_{1}^{(K_{\text{max}})}, T_{2}^{(K_{\text{max}})}, \ldots, T_{N_{\text{max}}}^{(K_{\text{max}})}]    
\end{equation}


Here, \(N_k\) represents the total number of trees on a specific client, and \(K_{max}\) is the total number of clients. The client-specific data \(X^{(k)}\) is propagated through \(M_{\text{global}}\) to derive a federated global affinity matrix. For each tree \(T_{i}^{(k)}\) in \(M_{\text{global}}\), the local data \(X^{(k)}\) traverses the tree to identify the leaf nodes. The binary matrix \(C_{i}^{(k)}\) is then updated based on the occurrences of pairs of samples in the same leaf node. The global federated count matrix \(C_{\text{global}}\) is the sum of binary count matrices across all trees in \(M_{\text{global}}\):
\begin{equation}
C_{\text{global}} = \sum_{k=1}^{K_{\text{max}}} \sum_{i=1}^{N_{k}} C_{i}^{(k)} 
\end{equation}
The federated affinity matrix $\hat{A}_{\text{global}}$ is again obtained by normalizing the count matrix $C_{\text{global}}$ by its maximum value. 
\begin{equation}
\hat{A}_{\text{gobal}} = \frac{C_{\text{global}}}{\max(C_{\text{global}})}     
\end{equation}
The federated affinity matrix $\hat{A}_{\text{global}}$ can be clustered by any distance-based clustering method.

\subsection{Cluster-specific feature importance}
Once an unsupervised random forest is trained and a cluster solution is determined, the unsupervised random forest can be used as a predictive model and the full capacity to quantify feature importance can be exploited. To this end it is possible to derive cluster-specific feature importance in a one versus all manner given the cluster solution as the outcome class. Note, training in this case is still unsupervised, because an unsupervised splitting rule is exploited. The outcome classes are just used to assess feature importance by labeling the samples within the nodes.  

\section{Evaluation strategy}\label{sec:evaluation}
\subsection{Evaluation on synthetic data}

The unsupervised random forest featuring a novel splitting rule, here denoted as \textit{uRF}, was assessed for its ability to generate an affinity matrix suitable for clustering, and it was compared to Euclidean distance, a widely used, intuitive, and computationally effective method for generating affinity matrices. Euclidean distance is effective in discerning isolated compact globular structures by assuming that sample points are distributed around the mean, but presents notable drawbacks, including sensitivity to scale and vulnerability to outliers. Synthetic datasets were generated to evaluate the clustering performance of the Ward linkage method on affinity matrices derived from the proposed method, Euclidean distance on non-normalized data, and Euclidean distance on standard normalized data. 
This evaluation encompasses scenarios where Euclidean distance is expected to perform well and those where it might face challenges.
Four sets of experiments were conducted to assess clustering performance in diverse scenarios: globular clusters of equal size, globular clusters with outliers, globular clusters of varying sizes, and non-globular clusters shaped as concentric circles. 30 synthetic datasets of two features were generated for each scenario and parameter configuration. The unsupervised random forests were trained with 500 trees, leaf node size set to 5, and one feature uniformly sampled from the feature space at each node.
The performance of Ward clustering on affinity matrices obtained using the three methods was quantified using the Adjusted Rand Index (ARI). 

In the first set of experiments, synthetic datasets were generated with three globular clusters of varying distances to evaluate the method with respect to the degree of separation among clusters. Each cluster comprises 100 data points distributed normally around centers located at coordinates (1, 0), (0, 1), and (1, 1). The degree of separation is controlled by a standard deviation ranging from 0.1 to 0.5 at intervals of 0.1, resulting in improved separation as the standard deviation decreases.
Then, synthetic datasets with globular clusters were generated with a varying percentage of outliers ranging from 2\% to 10\% at intervals of 2\%. Here, cluster data are positioned around the same cluster centers as in the previous case with a constant standard deviation of 0.25, while outliers are generated with centers in (3, 0), (0, 3), and (3, 3) and standard deviation of 1.
Next, synthetic datasets with three globular clusters of different sizes and varying distances among clusters were generated with centers at coordinates (0, 0), (1, 1), and (-2, 2) and standard deviations equal to 0.1, 0.1 + $m$0.1, and 0.1 + $m$0.2, with $m$ ranging from 1 to 5, so that as $m$ decreases, the clusters become better separated while maintaining different sizes. For example, for $m$ equal to 3, clusters have standard deviations equal to 0.1, 0.4, and 0.7, with a difference in standard deviation between each cluster and the next in size of 0.3.
The final set of experiments examined the performance of the clustering methods on non-globular clusters, specifically two concentric rings centered at coordinates (0,0). The data for the first ring was simulated with a minimum and maximum radius of 1 and 2, respectively, resulting in a width of 1. The second ring, also with a width of 1, was generated at a distance from the first ring ranging from 1 to 3 at intervals of 0.5.

\subsection{Sanity checks on machine learning benchmark datasets}
To evaluate the herein proposed unsupervised splitting rule we conducted experiments on six machine learning benchmark data sets (Table \ref{tab:Datasets}), which all are annotated with classification labels. 
We generated an affinity matrix utilizing the proposed unsupervised random forest which we used for Ward's linkage clustering method. We conducted a comparison with results obtained by calculating the affinity matrix using the Euclidean distance, once based on non-normalized data and another time with standard normalized data. For the random forest the default leaf node size was set to $5$, the number of features sampled from the feature space was set to $2$, and the number of trees was $100$. We varied the number of features for building a decision tree as well as the minimum leaf size. However, the fact that only two features are sampled from the feature space results in an increased randomization in the tree-building process. Recent work on unsupervised random forests highlight the beneficial behavior of this setting when the goal is to derive an affinity matrix for unsupervised clustering \cite{bicego2021learning}. 
As a performance metric accounting for cluster quality we used ARI. 

\begin{table*}[h!]
  \begin{center}
    \caption{Benchmark Datasets}
    \label{tab:Datasets}
    \begin{tabular}{l c c c c} 
     &  &  &  & \#Samples per \\ 
      
      \textbf{Datasets} & \#Samples & \#Features & \#Cluster & Cluster \\ 
      \hline
      Breast cancer &  106 & 9 & 6 & 21,15,18,16,14,22\\
      Glass    & 214 & 9 & 4  & 70,76,17,51\\
      Ionosphere & 351 & 34 & 2 & 225,126 \\ 
      Iris & 150 & 4 & 3  & 50,50,50\\
      Parkinson & 195 & 22 & 2& 48,147 \\
      Wine & 178 & 13 & 3 & 59,71,48 \\
      \hline
    \end{tabular}
  \end{center}
  \centering
\end{table*}

\subsection{Federated patient stratification based on multi-omics data}
Before we evaluated the proposed methodology in a federated setting, we investigated the general capacity of our method to detect disease subtypes. For this purpose we retrieved cancer data from The Cancer Genome Atlas (TCGA). Specifically, we retrieved data from four different cancer types, namely glioblastoma multiforme (GBM), kidney renal clear cell carcinoma (KIRC), sarcoma (SARC), and acute myeloid leukemia (AML). The data was pre-processed as follows: patients and features with more than $20\%$ missing values were removed and the remaining missing values were imputed with $k$-nearest neighbor imputation. In the Methylation data, we selected those 5000 features with maximal variance in each data set. All features were then normalised to have mean zero and a standard deviation of one. We randomly sampled 100 patients 30 times from the data pool and performed the Cox log-rank test, which is an inferential procedure for the comparison of event time distributions among independent (i.e. clustered) patient groups. 
We compared the results with SNF \cite{wang2014similarity}, NEMO \cite{rappoport2019nemo}, HCfused \cite{pfeifer2021hierarchical}, PINSplus \cite{nguyen2018pinsplus}, and Parea \cite{pfeifer2023parea}. 

In the federated case, we also randomly sampled 100 patients but distributed the subsampled data across three clients. Each client derived an affinity matrix on its local data and applied hierarchical clustering to obtain the clusters. After training, the local models were shared and the ensembles were concatenated to a global federated model. The local data was client-wise propagated through the global model to obtain an affinity matrix which also served as an input for Ward's linkage clustering method. We compared the clustering solution and quality from the local model with the global model. The aforementioned procedure was repeated 50 times. 

In an additional experiment we randomly sampled 100 patients, based on which a clustering solution was determined and then defined as the ground-truth. Subsequently, we distributed the data across three clients and we report on the performance of the local model compared to the performance of the global model. The procedure was repeated 50 times and we utilized ARI as a performance measure.

\section{Results and discussion}
\label{sec:results}

The evaluation of the proposed approach using synthetic datasets across diverse scenarios revealed that the unsupervised random forest with the novel splitting rule can generate affinity matrices that accurately reflect sample distances, leading to improved clustering when processed by algorithms such as Ward. Notably, the proposed method demonstrated comparable performance to Euclidean-based methods in scenarios where Euclidean distance is most effective, while it outperformed it significantly in other scenarios.

\begin{figure}[t]
    \centering
    \includegraphics[width=0.5\textwidth]{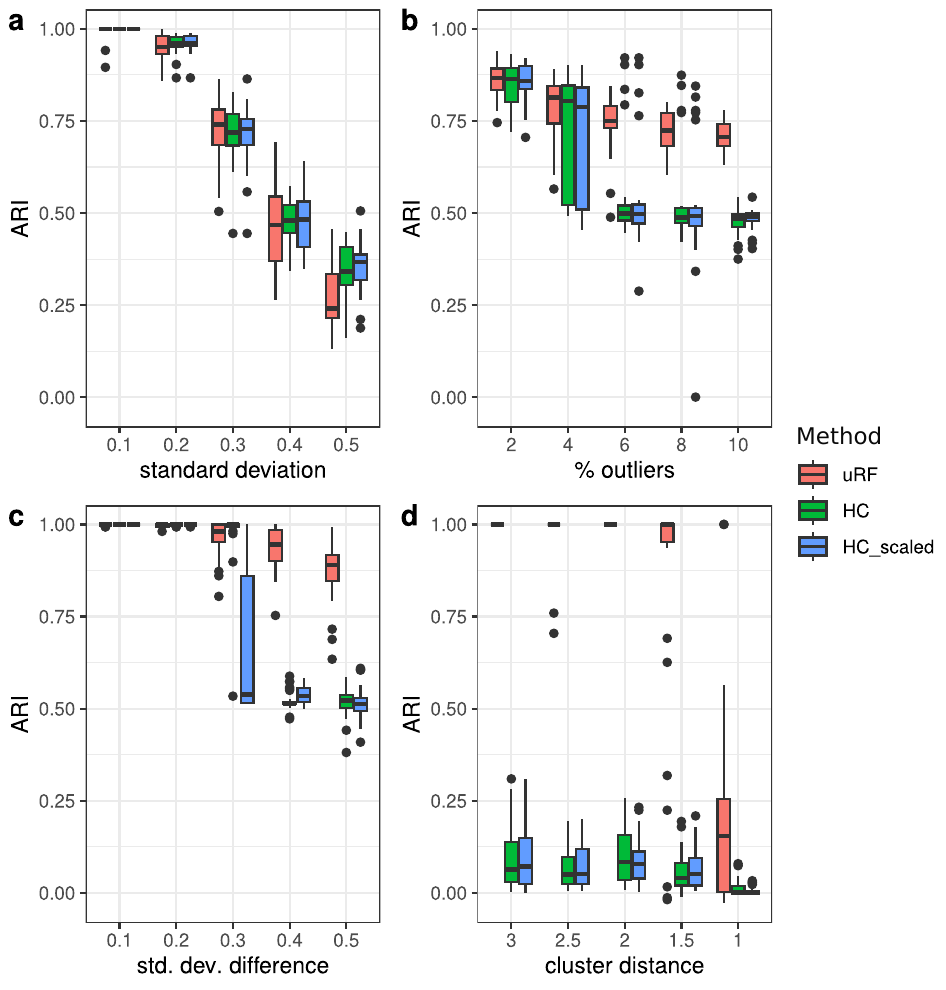}
    \caption{Clustering performance in terms of Adjusted Rand Index (ARI) of the Ward clustering algorithm on affinity matrices derived from the proposed unsupervised Random Forest with novel split rule \textit{uRF}, Euclidean distance (HC) and Euclidean distance on standardised data (HCscaled), evaluated in four scenarios: (a) globular clusters of equal size, (b) globular clusters with outliers, (c) globular clusters of varying sizes, and (d) non-globular clusters shaped as concentric circles.}
    \label{fig:synthetic}
\end{figure}

In the case of globular clusters of the same size, a scenario where Euclidean distance is expected to perform well, all three methods experience an expected decrease in ARI as cluster overlap increases. However, their performance remains consistently comparable across all experiments, as illustrated in Figure \ref{fig:synthetic}a. Further experiments conducted on well-separated globular clusters with different densities, obtained by assigning different numbers of samples to each cluster, revealed that all three methods are also robust to varying density and consistently deliver comparable results.
In the presence of outliers, all methods experience a decline in accuracy as the percentage of outliers increases, as shown in Figure \ref{fig:synthetic}b. Remarkably, \textit{uRF} maintains stable performance above 0.7, whereas Euclidean-based methods drop below 0.5 for outlier percentages exceeding 4\%. In the case of clusters of different sizes, shown in Figure \ref{fig:synthetic}c, \textit{uRF} maintains a good performance with an ARI consistently above 0.85, while other methods can effectively distinguish clusters only when well-separated, with performance dropping to 0.5 as clusters overlap. 
The ability to distinguish clusters of varying sizes is particularly relevant in the context of sub-clusters. Additional experiments, where synthetic data was generated to include one main cluster and two sub-clusters, yielded similar outcomes: Euclidean-based methods are only effective when clusters are sufficiently separated while \textit{uRF} consistently demonstrates good performance in all cases.
While Euclidean-based methods enable effective clustering for globular data provided that there is sufficient separation among clusters, this does not extend to non-globular clusters, as the method cannot detect non-convex structures. The results in Figure \ref{fig:synthetic}d show that regardless of the distance between the two rings, these methods struggle to accurately identify the clusters, yielding an ARI below 0.1. In contrast, \textit{uRF} can effectively discriminate the clusters when they are sufficiently separated, achieving an ARI of 1 for distances above 1.5. Additional experiments with other non-convex cluster shapes, such as half moons, confirmed these observations.

These results underscore the limitations of Euclidean distance in various clustering scenarios, highlighting the proposed approach's flexibility and robust performance in the presence of outliers, varying cluster sizes, and non-convex shapes, commonly encountered in omics data analysis.

\begin{figure}
     \centering
     \begin{subfigure}[b]{0.50\textwidth}
         \centering
         \includegraphics[width=\textwidth]{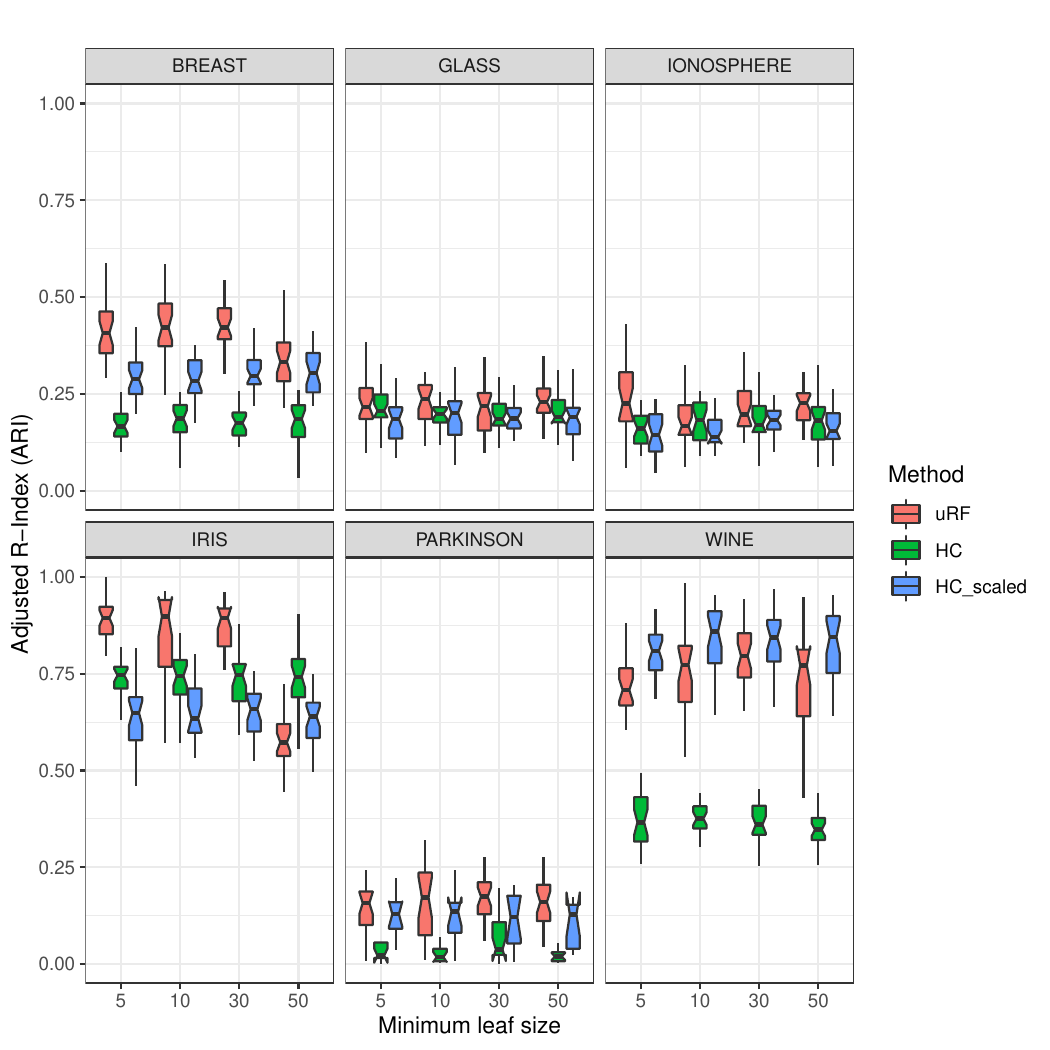}
         \caption{Varying the minimum leaf size.}
     \end{subfigure}
     \begin{subfigure}[b]{0.50\textwidth}
         \centering
         \includegraphics[width=\textwidth]{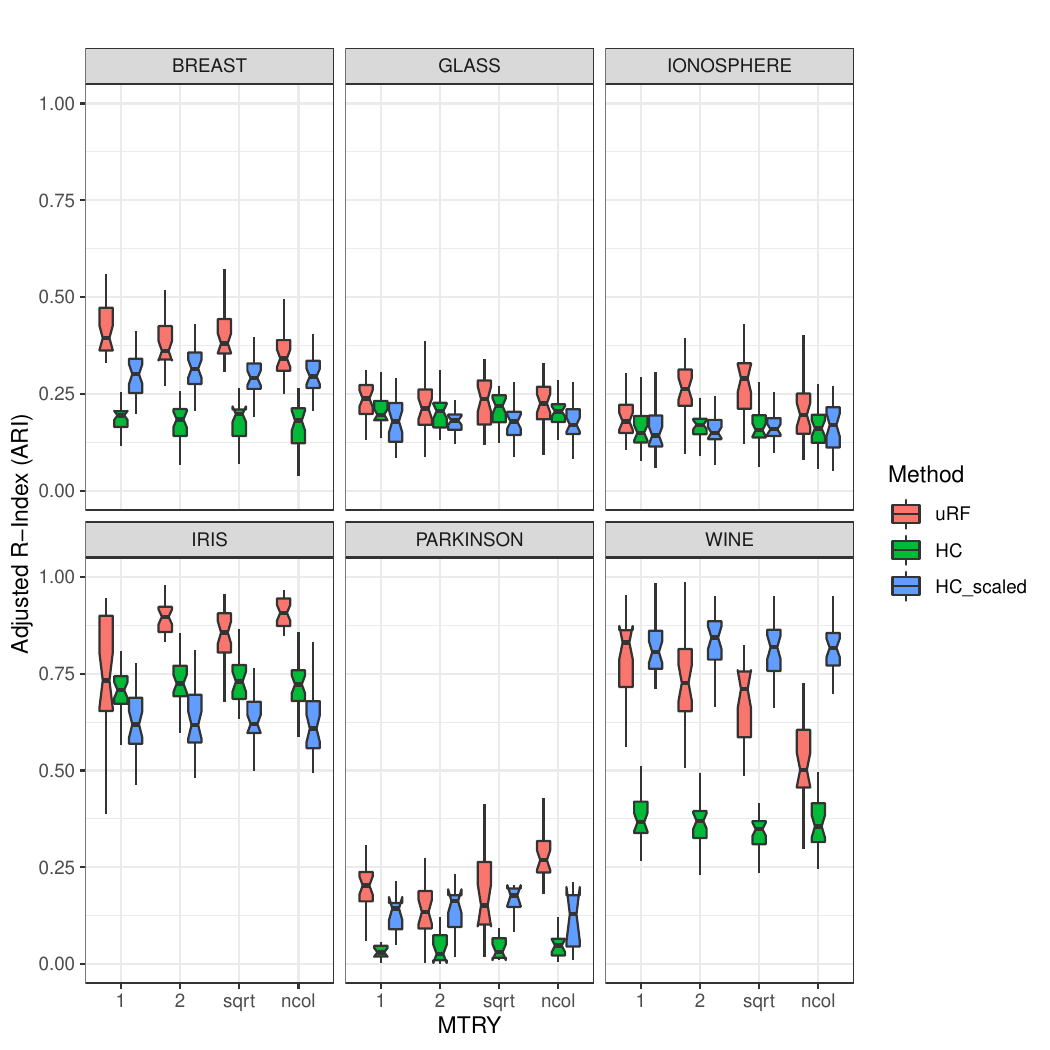}
         \caption{Varying the number of sampled features.}
     \end{subfigure}
        \caption{ (a) Varying the minimum size of leaf nodes. (b) Varying the number of sampled features. The resulting dendrograms are cut such that the number of clusters align with the ground-truth.} 
        
        \label{fig:Benchmark}
\end{figure}

The random forest-derived affinity matrix using our novel splitting rule was beneficial across almost all benchmark data sets (see Figure \ref{fig:Benchmark}a). 
In five out of six cases we obtained superior results compared to the Euclidean distance-derived affinity matrix. Furthermore, we could observe that the  minimum leaf size is a crucial parameter. In the case of Iris, for instance, a minimum leaf node size of 50 drastically decreases the clustering performance. The breast cancer data set was also sensitive to this specific parameter. 


%

The number of features available for splitting a node also had a considerable effect on the outcome. The square root of the total number of available features is often specified as a default. However, from Figure \ref{fig:Benchmark}b we can see that for the Breast cancer data set, as well as for the Wine data this might not be the best possible choice. In case of the Wine data set, for instance, an increasing number of available candidate features has a negative effect on the accuracy.
Overall, given the results in Figure \ref{fig:Benchmark} we could claim that the proposed unsupervised splitting rule serves our clustering purposes.

\begin{figure}
\centering
         \includegraphics[scale=.49]{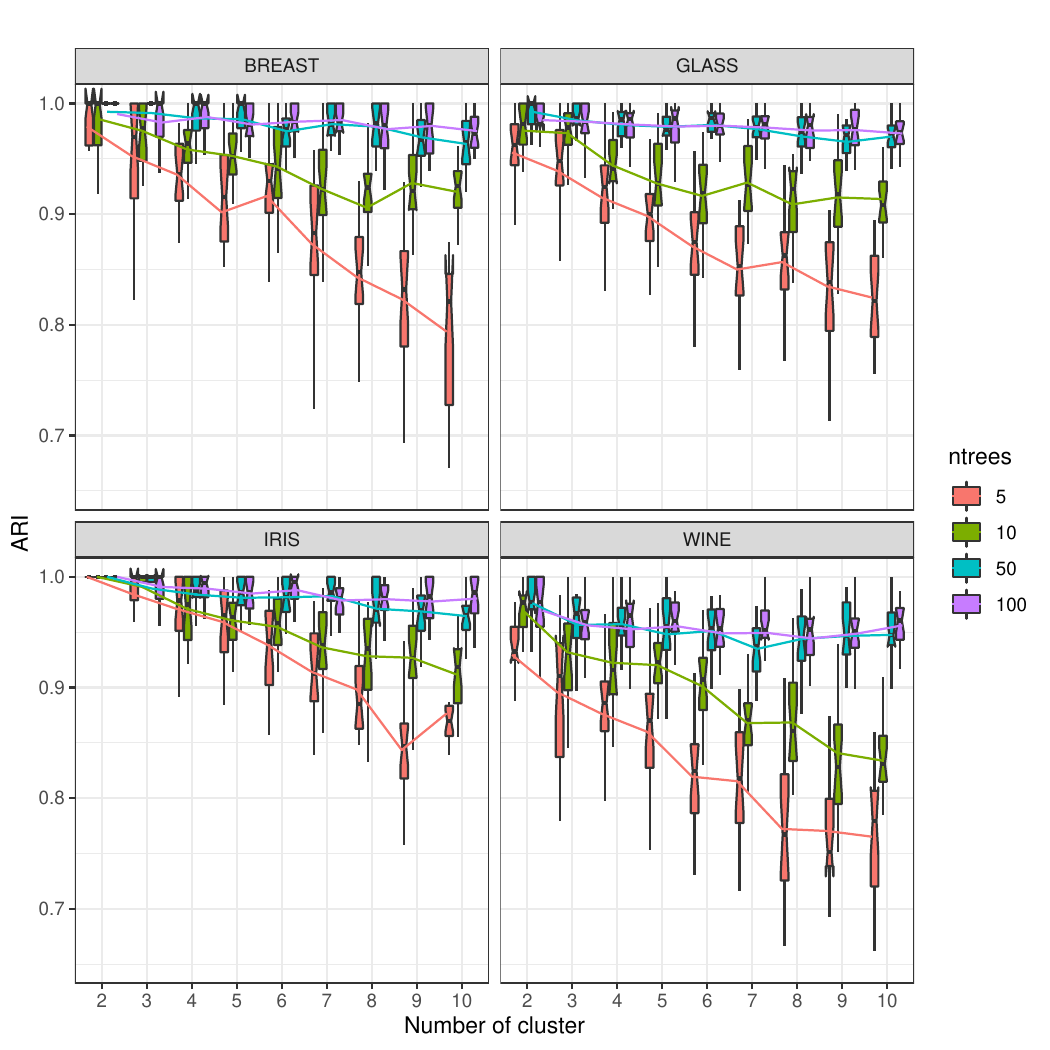}
         \caption{Verifying the optimal number of clusters using the proposed unsupervised random forest by subsequently reducing the number of trees. The clustering solutions at varies levels of $k$ were created using an unsupervised random forest comprising 500 trees. The derived affinity matrix served as an input for hierarchical clustering. The dendrogram was cut at different $k$ levels and the resulting multi-class labels were passed back to the unsupervised random forest as a response vector. In this way we label the samples within the leaf nodes of the unsupervised random forest to allow for predictions.}
         \label{fig:best_k}
\end{figure}

Moreover, we investigated a model-internal approach to determine the optimal number of clusters (see Figure \ref{fig:best_k}). The underlying assumption was that the best $k$ value is the most stable solution when the number of trees are decreased. We first derived an affinity matrix using the herein proposed unsupervised random forest. Afterwards, we created a dendrogram using Ward's hierarchical clustering. We cut the tree at a level where it resulted in a specified number of clusters $k$. From this clustering solution we created a multi-class response vector. These labels served as an input for the unsupervised random forest classifier to label the samples. With this approach we have created a predictive machine learning model. We checked the predictive performance of this model using the ARI index. Overall, we could learn that the classifier does reflect Ward's clustering solution very well, as indicated by the high ARI values (see Figure \ref{fig:best_k}). We subsequently reduced the number trees for a given $k$ and we tracked the ARI performance accordingly. A visual inspection of the plots in Figure~\ref{fig:best_k} gives us a hint what the optimal $k$ might be. In the case of Iris, for instance, a decrease in the number of trees has almost no effect at the $k=2$ level, thus we can assume that at least two clusters are present within the data. When $k=3$ is applied, the median ARI value is still at its maximum with $ARI=1$. Starting with $k=4$, there is a visible drop in performance when the number of trees are decreased. This observation suggest $k=3$ clusters within the data, which aligns perfectly with the ground-truth. 

\begin{figure}
\centering
         \includegraphics[scale=.55]{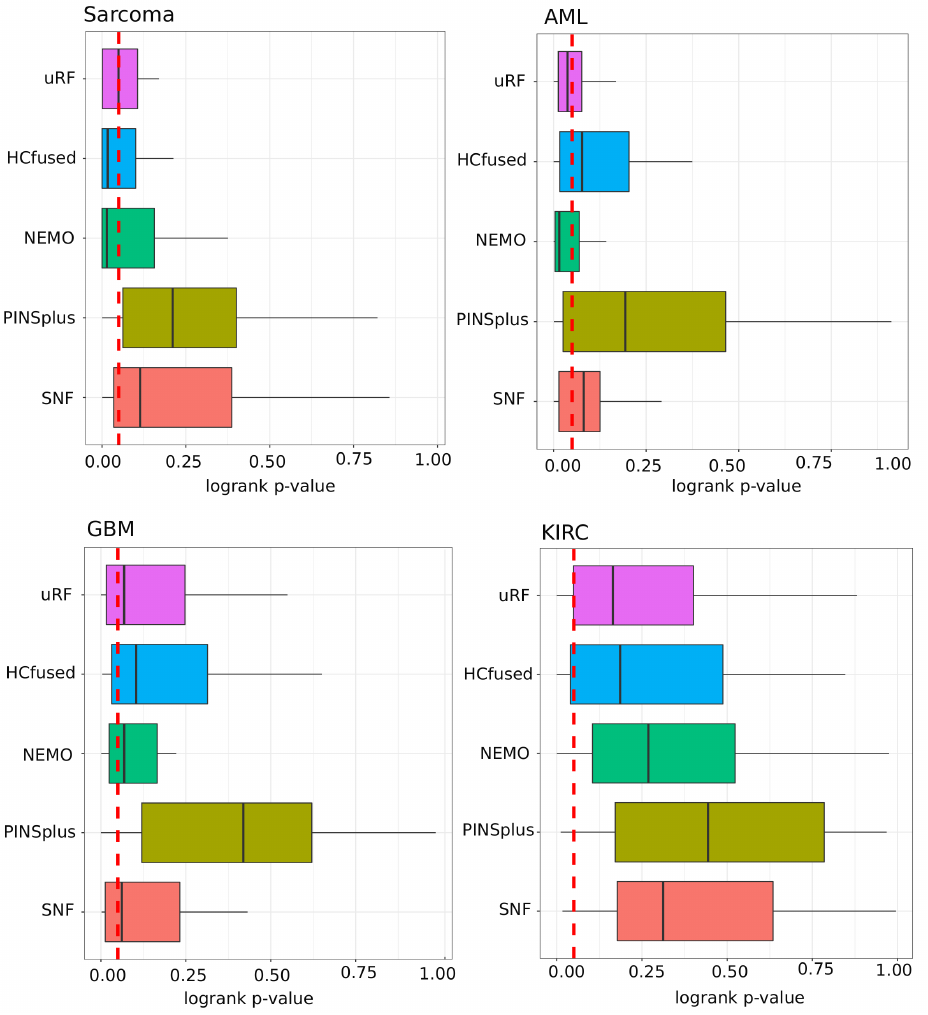}
         \caption{Non-federated disease subtype discovery based on multi-omics data in comparison with alternative approaches. Coloured bars represent the method-specific $p$-values of the Cox log-rank test from 30 iterations. The vertical line refers to the $\alpha= 0.05$ significance level. In case of \textit{uRF} the Silhouette coefficient was used to determine the optimal number of clusters.}
          \label{fig:TCGA}
\end{figure}

\begin{figure*}
\centering
         \includegraphics[scale=.65]{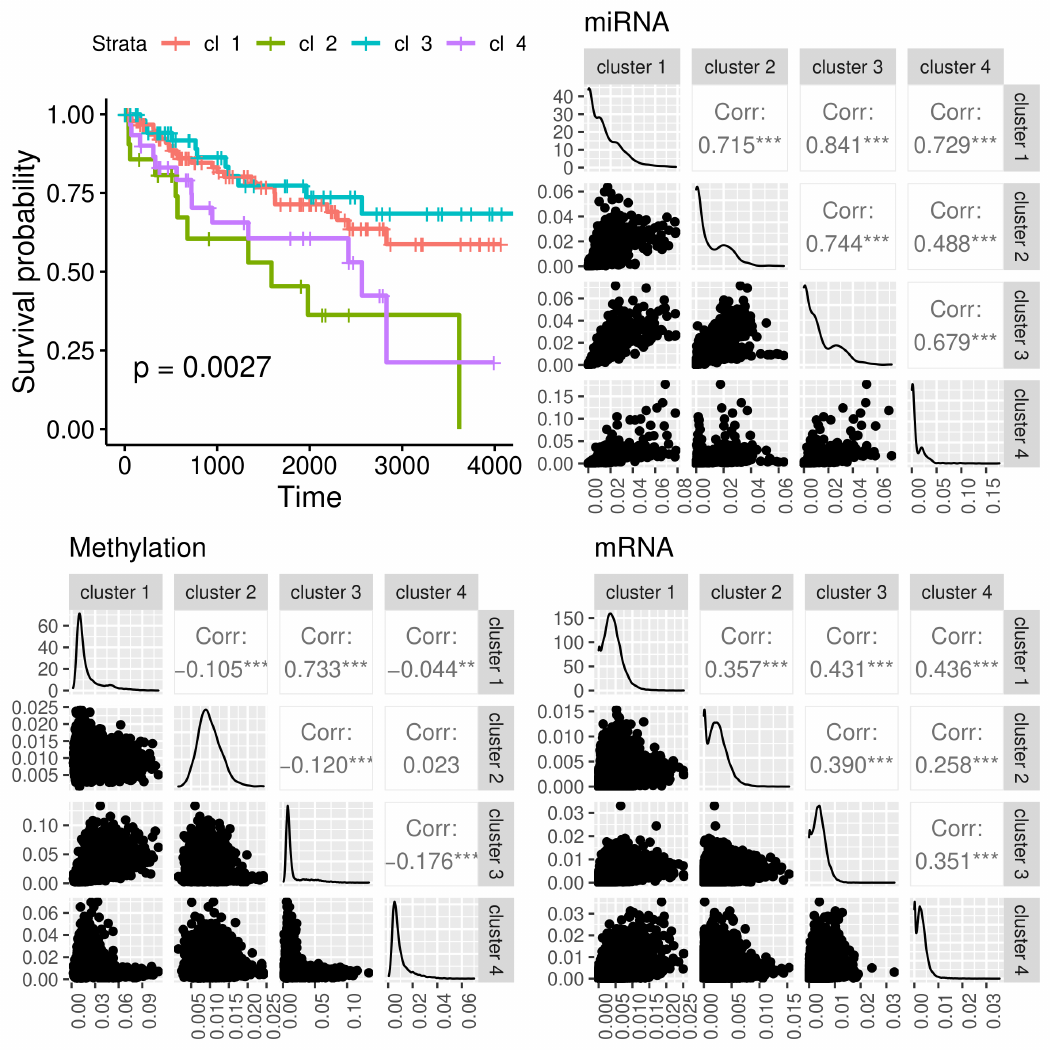}
         \caption{Kidney (KIRC) cancer data set. Survival curves of the four detected clusters are displayed. The other panels show the cluster-specific feature importance values and their inter-cluster correlation.}
         \label{fig:FI}
\end{figure*}

In the application on cancer data we do not have the ground-truth classification labels. While our model-internal approach to infer the best number of clusters is promising, it still cannot be executed in an automatic way, and therefore it is more suitable to perform a detailed visual inspection. Thus, we decided to utilize the Silhouette coefficient to infer the optimal number of clusters. The Silhouette coefficient has been shown to work well for disease subtyping purposes \cite{pfeifer2021hierarchical, pfeifer2023parea}. From Figure \ref{fig:TCGA} we can see that our approach, here denoted as \textit{uRF} 
, is competitive with the state-of-the-art. Our \textit{uRF} method clearly outperforms PINSplus and SNF, and is competitive with NEMO and HCfused. For Sarcoma and AML the median is below the $\alpha=0.05$ significance level. Given these results we could claim that the derived affinity matrix using our unsupervised random forest is suitable for omics-based patient stratification.

We further illustrate the versatility of our method on the kidney cancer (KIRC) data set. In Figure \ref{fig:FI} we show the survival curves based on a clustering on the whole data set, without any subsampling (as it is done in Figure \ref{fig:TCGA}). Our algorithm inferred four clusters. Cluster 2 and cluster 4 are high risk clusters, where a high mortality can be observed. We also show the cluster-specific feature importance and the Pearson correlation of these importance values between the detected clusters. The correlation plots suggest that cluster 2 and cluster 4 differ substantially in the Methylation data. Cluster 1 and cluster 3 have a high correlation of $r=0.73$, whereas all other cluster combinations are almost uncorrelated. A detailed inspection of the displayed data points would allow us to detect the genes or markers with the greatest signal differences.     

\begin{figure}
\centering
         \includegraphics[scale=.50]{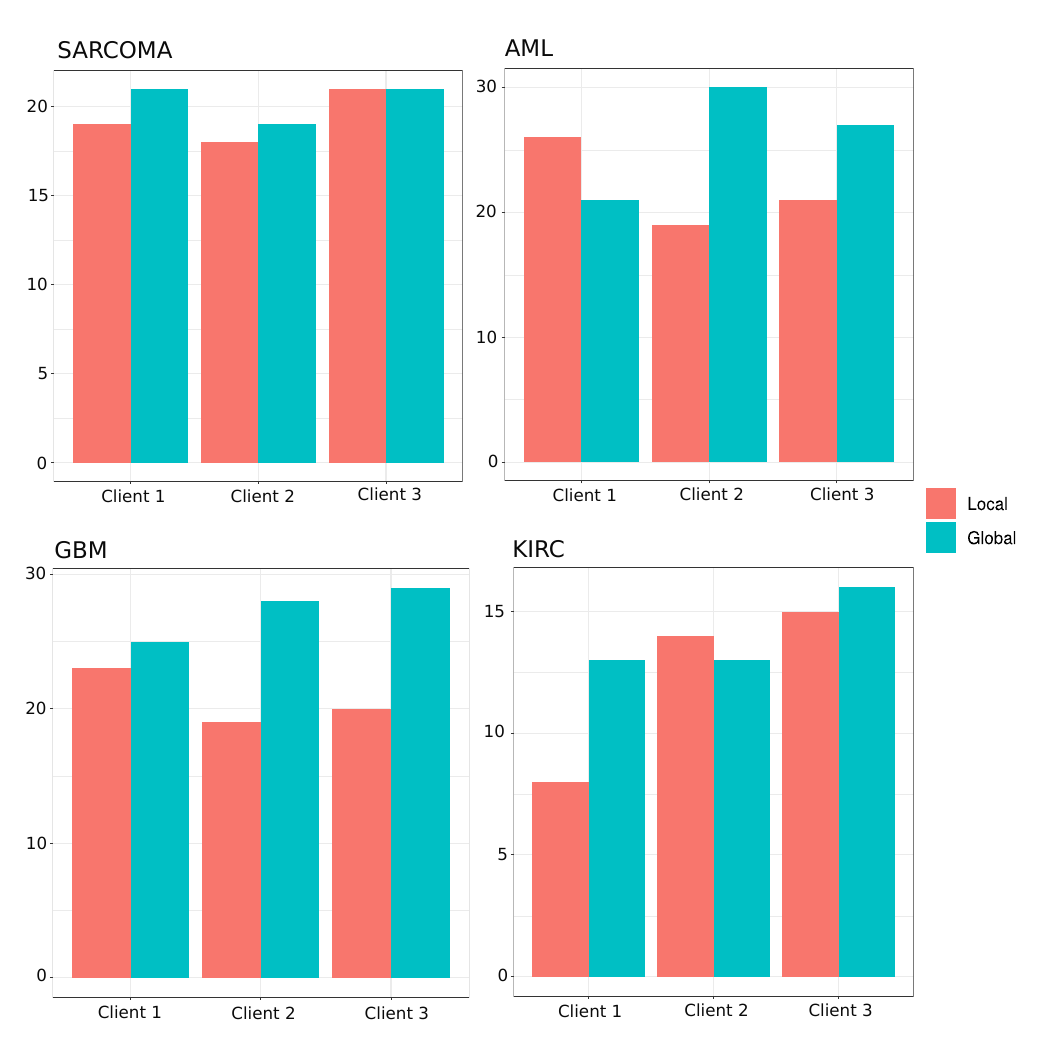}
         \caption{\textbf{Federated disease subtype discovery based on multi-omics data.} Shown is the number of times the global model outperformed the local model and vice-versa, on four different cancer types. The results are based on 50 iterations where we randomly distributed the data across three clients. The results are compared using the $p$-values of the Cox log-rank test. 
         }
         \label{fig:TCGA_fed}
\end{figure}

\begin{figure}
\centering
         \includegraphics[scale=.50]{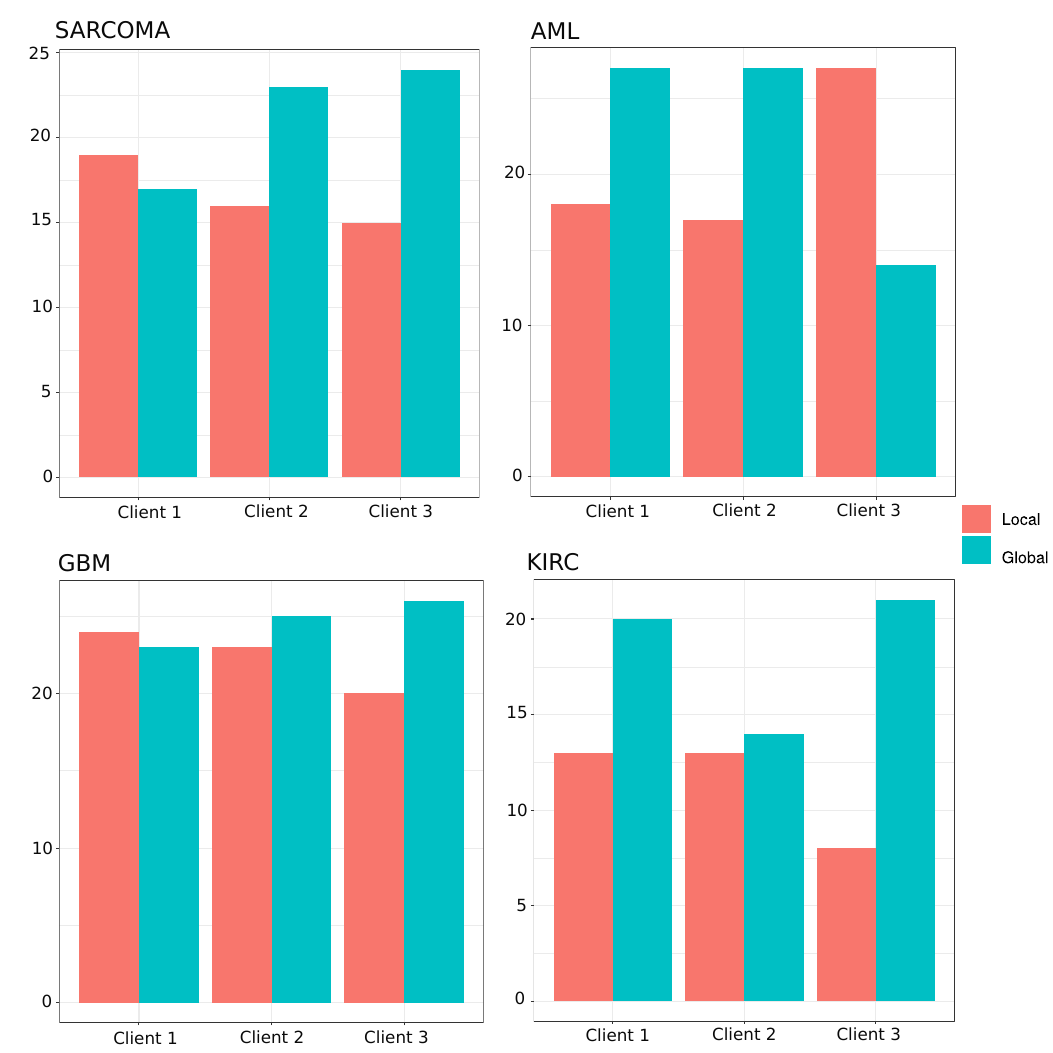}
         \caption{\textbf{Federated multi-omics clustering.} Shown is the number of times the global model outperformed the local model and vice-versa, on four different cancer types. The results are based on 50 iterations where we randomly distributed the data across three clients. The results are compared using the ARI index. 
         }
         \label{fig:TCGA_fed2}
\end{figure}

Finally, in the federated experiment, we could show that the aggregated global ensemble model performs well, and often better, than the local model. In all cases, the clustering quality verified by the Cox log-rank $p$-value was higher for at least two clients (see Figure \ref{fig:TCGA_fed}) and not far behind for the remaining cases. The results from the second experiment, where we first calculate the ground-truth on the whole data set and afterwards distribute the data, shows almost identical results (see Figure \ref{fig:TCGA_fed2}) to what we report in Figure \ref{fig:TCGA_fed}. From this experiment we can learn that a federated random forest subsequently trained on multiple subsets of the data is competitive with a model which was trained on the whole data set. It should be noted, however, that in the case of AML the local model from client 3 performed substantially better than the global model (see Figure \ref{fig:TCGA_fed2}).

Overall, these findings emphasize the potential of federated learning in achieving comparable results to centralized approaches while preserving data privacy and distribution across multiple clients.



\section{Conclusion}\label{sec:conclusion}
In conclusion, our study introduces a pioneering approach to federated patient stratification, leveraging a novel unsupervised splitting rule to derive an affinity matrix. The subsequent application of hierarchical clustering on this matrix has demonstrated promising results on synthetic data, machine learning benchmark datasets, and real-world cancer data. Our approach not only showcases robust performance in accurately categorizing patients into meaningful subgroups but also addresses the challenges of privacy and data security using federated machine learning. Moreover, it enables the identification of feature importance specific to clusters, revealing the driving molecular factors influencing distinct patient groups.


\section*{Competing interests}
No competing interest is declared.

\section*{Author contributions statement}
B.P. developed the methods. B.P and C.S conceived and conducted the experiments. B.P., C.S, and M.U. analyzed the results. B.P, C.S, M.B, M.K, and M.U wrote the manuscript. All authors reviewed the manuscript.

\section*{Acknowledgments}
We would like to thank Klaus-Martin Simonic for helpful discussions.

\bibliographystyle{IEEEtran}
\bibliography{IEEEabrv,Bibliography}

\begin{thebibliography}{10}
\providecommand{\url}[1]{#1}
\csname url@rmstyle\endcsname
\providecommand{\newblock}{\relax}
\providecommand{\bibinfo}[2]{#2}
\providecommand\BIBentrySTDinterwordspacing{\spaceskip=0pt\relax}
\providecommand\BIBentryALTinterwordstretchfactor{4}
\providecommand\BIBentryALTinterwordspacing{\spaceskip=\fontdimen2\font plus
\BIBentryALTinterwordstretchfactor\fontdimen3\font minus \fontdimen4\font\relax}
\providecommand\BIBforeignlanguage[2]{{%
\expandafter\ifx\csname l@#1\endcsname\relax
\typeout{** WARNING: IEEEtran.bst: No hyphenation pattern has been}%
\typeout{** loaded for the language `#1'. Using the pattern for}%
\typeout{** the default language instead.}%
\else
\language=\csname l@#1\endcsname
\fi
#2}}

\bibitem{brauneck2023federated}
A.~Brauneck, L.~Schmalhorst, M.~M. Kazemi~Majdabadi, M.~Bakhtiari, U.~V{\"o}lker, C.~C. Saak, J.~Baumbach, L.~Baumbach, and G.~Buchholtz, ``Federated machine learning in data-protection-compliant research,'' \emph{Nature Machine Intelligence}, vol.~5, no.~1, pp. 2--4, 2023.

\bibitem{dayan2021federated}
I.~Dayan, H.~R. Roth, A.~Zhong, A.~Harouni, A.~Gentili, A.~Z. Abidin, A.~Liu, A.~B. Costa, B.~J. Wood, C.-S. Tsai, \emph{et~al.}, ``Federated learning for predicting clinical outcomes in patients with {COVID}-19,'' \emph{Nature Medicine}, vol.~27, no.~10, pp. 1735--1743, 2021.

\bibitem{leng2022benchmark}
D.~Leng, L.~Zheng, Y.~Wen, Y.~Zhang, L.~Wu, J.~Wang, M.~Wang, Z.~Zhang, S.~He, and X.~Bo, ``A benchmark study of deep learning-based multi-omics data fusion methods for cancer,'' \emph{Genome Biology}, vol.~23, no.~1, pp. 1--32, 2022.

\bibitem{lipkova2022artificial}
J.~Lipkova, R.~J. Chen, B.~Chen, M.~Y. Lu, M.~Barbieri, D.~Shao, A.~J. Vaidya, C.~Chen, L.~Zhuang, D.~F. Williamson, \emph{et~al.}, ``Artificial intelligence for multimodal data integration in oncology,'' \emph{Cancer Cell}, vol.~40, no.~10, pp. 1095--1110, 2022.

\bibitem{rappoport2018multi}
N.~Rappoport and R.~Shamir, ``Multi-omic and multi-view clustering algorithms: review and cancer benchmark,'' \emph{Nucleic Acids Research}, vol.~46, no.~20, pp. 10\,546--10\,562, 2018.

\bibitem{subramanian2020multi}
I.~Subramanian, S.~Verma, S.~Kumar, A.~Jere, and K.~Anamika, ``{Multi-omics Data Integration, Interpretation, and Its Application},'' \emph{Bioinformatics and Biology Insights}, vol.~14, pp. 1--24, 2020.

\bibitem{wang2014similarity}
B.~Wang, A.~M. Mezlini, F.~Demir, M.~Fiume, Z.~Tu, M.~Brudno, B.~Haibe-Kains, and A.~Goldenberg, ``Similarity network fusion for aggregating data types on a genomic scale,'' \emph{Nature Methods}, vol.~11, no.~3, pp. 333--337, 2014.

\bibitem{rappoport2019nemo}
N.~Rappoport and R.~Shamir, ``{NEMO}: cancer subtyping by integration of partial multi-omic data,'' \emph{Bioinformatics}, vol.~35, no.~18, pp. 3348--3356, 2019.

\bibitem{von2007tutorial}
U.~Von~Luxburg, ``A tutorial on spectral clustering,'' \emph{Statistics and Computing}, vol.~17, no.~4, pp. 395--416, 2007.

\bibitem{yang2021subtype}
H.~Yang, R.~Chen, D.~Li, and Z.~Wang, ``Subtype-{GAN}: a deep learning approach for integrative cancer subtyping of multi-omics data,'' \emph{Bioinformatics}, vol.~37, no.~16, pp. 2231--2237, 2021.

\bibitem{yang2023mrgcn}
B.~Yang, Y.~Yang, M.~Wang, and X.~Su, ``{MRGCN}: cancer subtyping with multi-reconstruction graph convolutional network using full and partial multi-omics dataset,'' \emph{Bioinformatics}, vol.~39, no.~6, p. btad353, 2023.

\bibitem{pfeifer2021hierarchical}
B.~Pfeifer and M.~G. Schimek, ``A hierarchical clustering and data fusion approach for disease subtype discovery,'' \emph{Journal of Biomedical Informatics}, vol. 113, p. 103636, 2021.

\bibitem{pfeifer2023parea}
B.~Pfeifer, M.~D. Bloice, and M.~G. Schimek, ``Parea: multi-view ensemble clustering for cancer subtype discovery,'' \emph{Journal of Biomedical Informatics}, vol. 143, p. 104406, 2023.

\bibitem{Breiman2001-rl}
L.~Breiman, ``{Random Forests},'' \emph{Machine Learning}, vol.~45, no.~1, pp. 5--32, 2001.

\bibitem{wright1949genetical}
S.~Wright, ``The genetical structure of populations,'' \emph{Annals of Eugenics}, vol.~15, no.~1, pp. 323--354, 1949.

\bibitem{holsinger2009genetics}
K.~E. Holsinger and B.~S. Weir, ``Genetics in geographically structured populations: defining, estimating and interpreting {FST},'' \emph{Nature Reviews Genetics}, vol.~10, no.~9, pp. 639--650, 2009.

\bibitem{ward1963hierarchical}
J.~H. Ward~Jr, ``{Hierarchical Grouping to Optimize an Objective Function},'' \emph{Journal of the American Statistical Association}, vol.~58, no. 301, pp. 236--244, 1963.

\bibitem{murtagh2014ward}
F.~Murtagh and P.~Legendre, ``{Ward’s hierarchical agglomerative clustering method: Which algorithms implement Ward’s criterion?}'' \emph{Journal of Classification}, vol.~31, no.~3, pp. 274--295, 2014.

\bibitem{hauschild2022federated}
A.-C. Hauschild, M.~Lemanczyk, J.~Matschinske, T.~Frisch, O.~Zolotareva, A.~Holzinger, J.~Baumbach, and D.~Heider, ``{Federated Random Forests can improve local performance of predictive models for various healthcare applications},'' \emph{Bioinformatics}, vol.~38, no.~8, pp. 2278--2286, 2022.

\bibitem{bicego2021learning}
M.~Bicego and F.~Escolano, ``{On learning Random Forests for Random Forest-clustering},'' in \emph{2020 25th International Conference on Pattern Recognition (ICPR)}.\hskip 1em plus 0.5em minus 0.4em\relax IEEE, 2021, pp. 3451--3458.

\bibitem{nguyen2018pinsplus}
H.~Nguyen, S.~Shrestha, S.~Draghici, and T.~Nguyen, ``{PINSPlus: a tool for tumor subtype discovery in integrated genomic data},'' \emph{Bioinformatics}, vol.~35, no.~16, pp. 2843--2846, 2019.

\end{thebibliography}

\vfill


\end{document}